**Large language models require a new form of oversight: capability-based monitoring**

Katherine C. Kellogg,[1] Bingyang Ye,[2,3,4] Yifan Hu,[5] Guergana K. Savova,[6] Byron Wallace,[7] Danielle S. Bitterman[2,3,6]

1. MIT Sloan School of Management, Boston, MA, USA
2. AI in Medicine Program, Mass General Brigham, Harvard Medical School, Boston, MA, USA
3. Department of Radiation Oncology, Brigham and Women's Hospital/Dana-Farber Cancer Institute, Boston, MA, USA
4. Department of Computer Science, Brandeis University, Waltham, MA, USA
5. Harvard John A. Paulson School Of Engineering And Applied Sciences, Cambridge, MA, USA
6. Computation Health Informatics Program, Boston Children's Hospital, Harvard Medical School, Boston, MA, USA
7. Khoury College of Computer Sciences, Northeastern University, Boston, MA, USA

**Corresponding Authors:**
Dr. Danielle S. Bitterman
AI in Medicine Program
Department of Radiation Oncology
Mass General Brigham
75 Francis Street, Boston, MA 02115

Katherine C. Kellogg
David J. McGrath jr (1959) Professor of Management and Innovation
Department of Work and Organization Studies
MIT Sloan School of Management
E62-324, 100 Main Street
Cambridge, MA 02142

**Disclosures:**

DSB: Associate Editor, JCO Clinical Cancer Informatics (not related to the submitted work), Associate editor of Radiation Oncology of HemOnc.Org (not related to the submitted work) and is on the Scientific Advisory Board of Mercurial AI (not related to the submitted work) and Blue Clay Health LLC (not related to the submitted work).

**Funding:**

The authors acknowledge financial support from the National Institutes of Health National Cancer Institute (U54CA274516-01A1 [G.K.S., D.S.B], R01CA294033-01 [B.Y., G.K.S., D.S.B], U24CA248010 [B.Y., G.K.S., D.S.B]), the American Cancer Society and American Society for Radiation Oncology, ASTRO-CSDG-24-1244514-01-CTPS Grant DOI #: https://doi.org/10.53354/ACS.ASTRO-CSDG-24-1244514-01-CTPS.pc.gr.222210 [G.K.S., D.S.B.], a Patient-Centered Outcomes Research Institute (PCORI) Project Program Award (ME-2024C2-37484) [D.S.B.], and the Woods Foundation [D.S.B.]. All statements in this report, including its findings and conclusions, are solely those of the authors and do not necessarily represent the views of the Patient-Centered Outcomes Research Institute (PCORI), its Board of Governors or Methodology Committee.

## Abstract

The rapid adoption of large language models (LLMs) in healthcare has been accompanied by scrutiny of their oversight. Existing monitoring approaches, inherited from traditional machine learning (ML), are task-based and founded on assumed performance degradation arising from dataset drift. In contrast, with LLMs, inevitable model degradation due to changes in populations compared to the training dataset cannot be assumed, because LLMs were not trained for any specific task in any given population. We therefore propose a new organizing principle guiding generalist LLM monitoring that is scalable and grounded in how these models are developed and used in practice: capability-based monitoring. Capability-based monitoring is motivated by the fact that LLMs are generalist systems whose overlapping internal capabilities are reused across numerous downstream tasks. Instead of evaluating each downstream task independently, this approach organizes monitoring around shared model capabilities, such as summarization, reasoning, translation, or safety guardrails, in order to enable cross-task detection of systemic weaknesses, long-tail errors, and emergent behaviors that task-based monitoring may miss. We describe considerations for developers, organizational leaders, and professional societies for implementing a capability-based monitoring approach. Ultimately, capability-based monitoring will provide a scalable foundation for safe, adaptive, and collaborative monitoring of LLMs and future generalist artificial intelligence models in healthcare.

The enthusiasm and rapid uptake of generalist artificial intelligence (AI) models, in particular large language models (LLMs), in healthcare has spurred much discussion of their evaluation and oversight for clinical applications. But less attention has been paid to the core assumptions about model performance degradation that underpin monitoring strategies, but that break down in the case of LLM use. Here, we propose a new capability-based monitoring framework that is better aligned with how LLMs are trained and used in practice, and describe implementation considerations for this novel approach.

Traditionally, AI implementations in healthcare have been focused on bespoke Machine Learning (ML) models trained for a single task using datasets from defined, bounded populations (Figure 1, ML Paradigm). These models assume that training and test data come from the same underlying distributions. When this assumption is violated, overfitting occurs, leading to degraded performance on new datasets.[1,2] ML models trained for a particular task, such as sepsis prediction,[3] on bounded, labeled clinical datasets reflecting (hopefully) their target clinical population, will always be to some extent overfit (that is, only performant for the task and populations they were trained on).[3,4] Because of this, performance degradation post-deployment is a given: models will always degrade because populations and outcome distributions inevitably change compared to data the model was trained on.[1,2] This has led, sensibly, to model-specific post-deployment monitoring for expected degradation.

In contrast, the emergence of generalist LLMs fundamentally challenges these prior assumptions driving performance monitoring. Despite not being trained using in-distribution clinical data or specifically for clinically-relevant tasks, LLMs can still capably summarize clinic visit transcripts into note drafts (ambient documentation),[5,6] answer clinical questions,[7] translate patient instructions,[8] and more. Inevitable model degradation due to changes in populations compared to the training dataset cannot be assumed, because LLMs were not trained for any specific task in any given population (Figure 1, LLM Paradigm). Many—probably most—clinical tasks will be “out-of-distribution” for the LLM training set, which is massive and often unknown anyway. Thus, traditional notions of ML overfitting and performance degradation do not straightforwardly apply. Performance variation due to LLM “overfitting” now occurs due to prompting, knowledge evolution, cultural shifts, and deployment environments, not training dataset distributions defined by input features and labels. While we shouldn’t anticipate degradation due to overfitting in the traditional sense, we do expect that an LLM will behave differently across populations in ways that are not necessarily predictable.

The generalist properties of LLMs make them powerful and drive uptake,[9] but also complicate monitoring. Accordingly, for LLMs and other similar generalist models, monitoring frameworks must evolve. Ongoing task-based oversight is not only impractical as LLMs drive task expansion, but also undesirable because it will leave us blind to shared vulnerabilities. We therefore propose a new organizing principle guiding generalist LLM monitoring that is grounded in how these models are developed and used in practice: capability-based monitoring.

Capability-based monitoring is motivated by the fact that LLMs are generalist systems whose overlapping internal capabilities are reused across numerous downstream tasks (Table 1). In this approach, tasks relying on the same underlying model and drawing on similar capabilities are monitored collectively—acknowledging that some tasks engage multiple capabilities simultaneously. For instance, the ability to summarize underlies a range of workflows with distinct contexts, such as inpatient discharge summary generation, outpatient pre-charting, and ambient documentation. Monitoring each task in isolation fragments oversight and risks missing cross-cutting vulnerabilities that propagate across tasks. In contrast, capability-based monitoring (Table 2) provides a more practical and comprehensive framework, enabling cross-task evaluation of shared operations, early detection of systemic weaknesses, and identification of edge cases or rare errors that task-specific monitoring might overlook (Figure 2). This is particularly critical for LLMs, which often struggle with infrequent but clinically significant long-tail scenarios.[7]

LLM performance degradation arises when models overfit the multi-dimensional contextual factors that shape their behavior. Intrinsic factors pertain to properties of the model itself, including its alignment with professional standards and values, temporal currency (i.e., how up-to-date its knowledge base is), reasoning quality, robustness to variation in input style or language, and compute efficiency.[10] Extrinsic factors involve human interaction, including the degree of human oversight and the type and extent of human–model collaboration, both of which impact overall system performance.[11,12] Table 2 outlines monitoring dimensions and proposed metrics encompassing these factors.

Not all dimensions currently validated have automatable monitoring approaches that are known to correlate with human evaluation; many still require human review and gold-standard comparators[13] although LLM evaluation strategies and metrics, including gaps specific to healthcare, have been extensively discussed in prior work.[13–18] Our framework aims to organize and prioritize metric development and validation. Existing benchmarks in both general and clinical domains, while imperfect, can also supplement real-world monitoring by identifying performance gaps within specific capabilities.[19–21]

Given the limited availability of validated metrics and ground truth labels they require, the LLM-as-judge paradigm (where a separate model is used to evaluate outputs) is gaining traction as a flexible, extensible monitoring method. We include LLM-as-judge as an automatable metric across several dimensions, but emphasize that these secondary models also require validation and ongoing oversight for each dimension in which they are applied (see Safety Guardrail Capability, Table 1).[22]

A monitoring strategy should not only identify errors, but lead to actionable corrections.[10] Importantly, performance degradation across dimensions does not always necessitate a full model update. Limitations arising from intrinsic factors may often be addressed through prompt refinement, improved tool integration, or adjustments to retrieval databases before modifying the underlying model. In contrast, extrinsic factors may call for interventions such as enhanced interface design, user training, or targeted education. We envision primary capability monitoring occurring on a per-LLM basis, as each model is trained on distinct datasets that are typically opaque to the institutions deploying them. Nevertheless, given the shared pretraining corpora and similar tuning paradigms among many LLMs, and the fact that it may not always be known when a vendor updates LLM's weights, vulnerabilities identified in one model should prompt systematic evaluation across related models.

As an example of the strengths of capability-based monitoring, envision an institution that has implemented 3 tasks requiring strong summarization capabilities: hospital course summarization, ambient documentation, and patient-facing discharge instructions (Figure 2). Missing information is flagged sparsely in all 3 tasks, and the signal only becomes significant when grouped, enabling identification that errors occur when input exceeds a context length threshold, thus a solution is implementation of a new preprocessing step to reduce context length. Similarly, a rare token (wordpiece) repeated many times in a single patient's inpatient notes is found to trigger biased language. Efficient simulations confirm this is a shared failure mode, so an input filter is implemented for all summarization tasks, avoiding future potential errors for all summarization workflows.

Implementing capability-based monitoring creates new challenges, implications, and benefits for healthcare organizations (Table 3). Key challenges for developers in healthcare organizations include: a) capability and monitoring dimensions are not yet fully scoped and taxonomized, and will increase over time, b) it is not feasible to manually monitor all of these metrics for all models, and c) human oversight and task-specific monitoring will likely still be required for very high risk applications. Developers can address these challenges by a) developing visualization-based dashboards and defining evaluation frequency and thresholds for error detection, b) engaging in two-tiered monitoring with automated screening by Judge LLMs and existing automated metrics (high-frequency, low-cost) and human review of flagged cases (low-frequency, high-interpretability), and c) experimenting with various techniques for addressing human automation bias, over-reliance, and de-skilling in order to provide effective human oversight of models deployed in high-risk tasks.

Key challenges for organizational leaders in healthcare organizations include: a) decentralized capability-based monitoring at the business unit-level gives business unit leaders more control, but risks missing cross-cutting vulnerabilities, b) merely detecting performance degradation through capability-based monitoring is not sufficient, c) individuals may develop LLM implementations via prompt refinement for private use and not report these to the organization for monitoring, and d) use of LLMs can deskill healthcare workers, making it difficult to take LLMs offline when deterioration is detected. Organizational leaders can address these challenges by a) centralizing capability-based monitoring while working with business unit stakeholders to identify and create specific data views and functionality required by these decentralized stakeholders, b) identifying who is accountable for diagnosing the root cause of degradation, and developing a set of methods for root cause diagnosis and restoring model performance, c) providing recognition, rewards and resources to individuals for formalizing new models, and d) instituting requirements that professionals regularly practice high-impact tasks without AI, to maintain proficiency.

While capability-based monitoring should enable more practical and robust oversight, future work is needed to realize its full potential at scale. First, our proposed capabilities and metrics are likely not exhaustive, and we encourage the community to contribute to a comprehensive taxonomy of each.

Although more streamlined than task-based monitoring, there are still many ways an organization may wish to visualize capability across models and business units.[23] Research into the optimal visualizations and interface for such monitoring tools will be needed to make sure they are usable and sustainable. Evaluation frequency and thresholds for error detection across all monitoring dimensions will need to be defined and refined as we gain experience implementing LLMs. The quality assurance, process improvement, and statistical quality control fields will play an important role in developing these thresholds. Finally, because many institutions will use the same underlying LLMs for various tasks, there is an enormous opportunity to extend this strategy to a collaborative monitoring commons across institutions. While capability-based collaborative monitoring will not require sharing data or models, it will require standardized documentation and logging of LLM use. Active uptake and expansion of efforts such as MedLog, a protocol for event-level clinical AI logging, will be critical in realizing this vision.[24]

In the LLM era, “overfitting” in healthcare AI has shifted from model training to prompt, context, and workflow over-adaptation, making the traditional distinction between in-distribution and out-of-distribution clinical data far less predictive of performance. Monitoring of generalist AI, exemplified by LLMs, should be fit-for-purpose: designed to address how LLMs are trained and used in practice, not simply extended from traditional models that have different performance and generalization assumptions. As such, the unit of monitoring must evolve from tasks to capabilities, tracking shared behaviors across contexts. Capability-based monitoring is at once technically necessary and organizationally scalable. Healthcare systems, vendors, and regulators should adopt capability-based frameworks to ensure safe, equitable, and sustainable deployment of generalist AI.

**Figure 1. Illustration of train and test data distributions in traditional Machine Learning (ML) models vs. Large Language Models.**

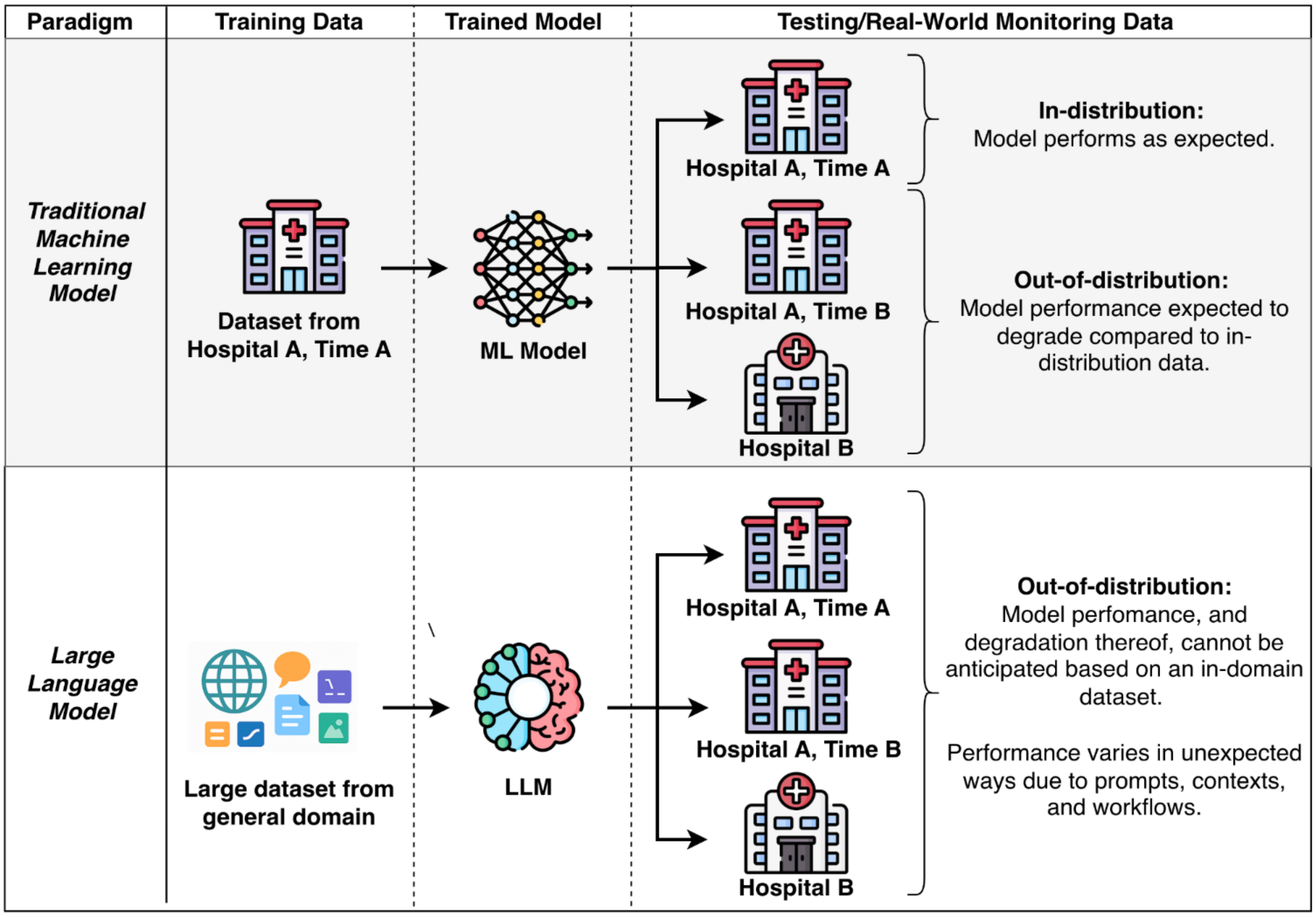

**Figure 1. Illustration of train and test data distributions in traditional Machine Learning (ML) models vs. Large Language Models (LLMs).** In traditional ML models, it is assumed that test data come from the same underlying distribution (i.e., in distribution; Hospital A, Time A in the figure). As models are applied to different real-world data distributions such as evolution over time (e.g., Hospital A, Time B) and/or new settings (e.g., Hospital B), performance optimized and reported on in-distribution data is no longer reliable. Instead, performance is anticipated to degrade due to overfitting. In LLMs, models are trained from large, general datasets and learn general abilities. All clinical datasets are out of distribution and traditional notions of ML overfitting and population drift due not straightforwardly apply.

**Figure 2. Aggregating Task-Level Signals via Capability-Based Monitoring Reveals Shared Failure Modes.**

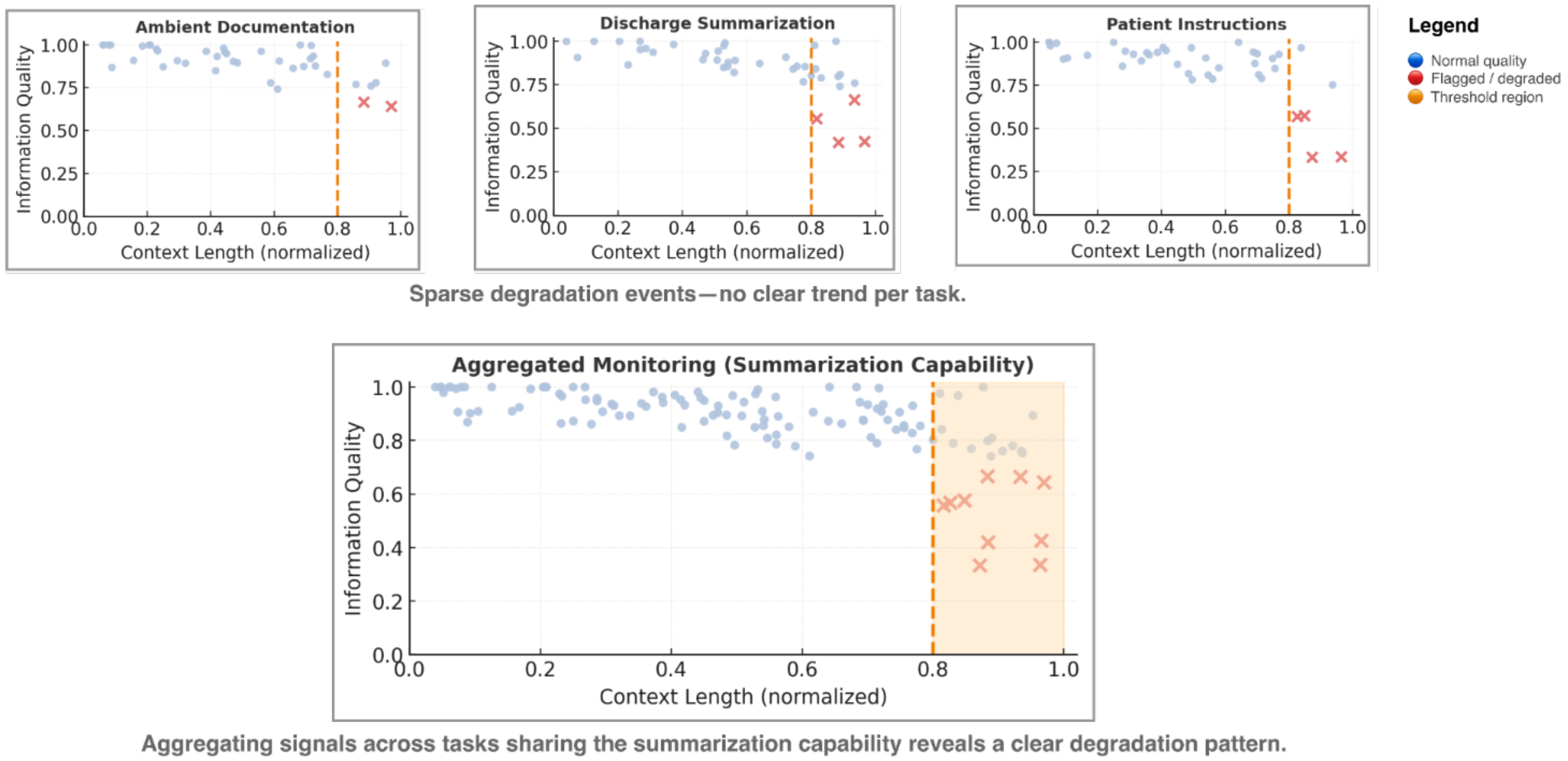

**Figure 2. Aggregating Task-Level Signals via Capability-Based Monitoring Reveals Shared Failure Modes.** Individual monitoring shows sparse quality issues across hospital course discharge summarization, ambient documentation, and patient instructions. Aggregated capability-level monitoring exposes a shared vulnerability when context length exceeds a threshold, enabling corrective preprocessing that restores performance across all summarization workflows.

**Table 1. Generalist Large Language Model Capability Families**

| Capabilities | Core capability | Examples |
|---|---|---|
| Summarization | Ability to compress text, preserve relevant facts, prioritize appropriate facts and documents | Hospital course summary generation for discharge notes, ambient documentation |
| Information Extraction | Identify and capture structured data from unstructured documentation Extract discrete fields, classification | Trial matching, medical coding, automated registry curation, medical coding |
| Classification | Assign predefined codes or categories to standardized concepts | Patient portal triage, outcome prediction, medical coding |
| Decision support/Question answering | Given inputs, provide appropriate clinical rationale or justification | Clinical decision support, insurance appeals, patient question-answering |
| Simplification | Generate patient-friendly language (lay and multi-lingual), reply to patients | Patient education, discharge instructions, machine translation |
| Translation | Multi-lingual translation and communication | Translation of patient-facing information, real-time translation in clinic |
| Information Retrieval | Identify and retrieve appropriate documents for an incoming query | Literature search, chart curation |
| Safety Guardrail* | Decide when to answer or block a response, safe completion, filter toxic language | Auditing models, hallucination detection, LLM-as-judge |

Abbreviations: LLM: Large Language Model

I

**Table 2. Monitoring dimensions and proposed metrics**

| Monitoring dimension | Description | Example metrics that require human labels | Example automated metrics | Priority monitoring dimensions for each capability family | | | | | | | |
|---|---|---|---|---|---|---|---|---|---|---|---|
| | | | | **Summarization** | **Information Extraction** | **Classification** | **Decision support/Question answering** | **Simplification** | **Translation** | **Information retrieval** | **Safety Guardrail** |
| Information quality and accuracy | Faithfulness to inputs, factual correctness | % factual errors in sampled outputs scored by human expert | LLM-as-judge<br>Medical knowledge benchmarks | X | X | X | X | X | X | X | X |
| Reasoning | Internal logic of output and clinical soundness | Human expert review | LLM-as-judge<br>Medical reasoning benchmarks | | | | X | | | X | |
| Style | Clarity, readability, tone appropriateness | Human expert review | Automated readability metrics<br>LLM-as-judge | X | | | X | X | X | | |
| Sycophancy and refusal behavior | Ability to decline unsafe or uncertain requests | Human expert review | LLM-as-judge<br>Sycophancy benchmarks | | | | X | | | X | X |
| Input robustness and feature drift | Stability under changing prompts, data quality | Human expert review | Overlap metrics<br>LLM-as-judge<br>Text-based statistics (e.g., input tokens) | X | X | X | X | X | X | X | X |
| Equity | Performance across subgroups (demographics, specialties) | N/A | Distribution of other metrics across subgroups, including raw distribution and | X | X | X | X | X | X | X | X |
| | | | fairness metrics | | | | | | | | |
| End-user preference | Human edit rates, acceptance ratios, escalation frequency | N/A | Edit distance, acceptance rate | X | X | X | X | X | X | X | |
| Toxicity | Presence of toxic, stigmatizing, or otherwise inappropriate language | Human expert review | LLM-as-judge Toxicity and bias benchmarks | X | | | X | X | X | X | |
| Process | Costs, energy, time | N/A | Tokens, costs, FLOPs, latency per unit time (to understand usage) and per query (to understand potential LLM behavior changes) | X | X | X | X | X | X | X | X |

Abbreviations: LLM: Large Language Model; FLOPS: Floating point operations per second

## Table 3. Monitoring Implementation and Oversight Design

| New Monitoring Challenges with LLMs | Implications for Practice | Benefits | Limitations | Specific Recommendations |
|---|---|---|---|---|
| **Implications for Developers** | | | | |
| Task-based monitoring fragments oversight and misses cross-cutting vulnerabilities | • Create registries/ dashboards that visualize performance metrics per capability, not per task | • Scalability and reduced redundancy in compliance and auditing | • Capability and monitoring dimensions not yet fully scoped and taxonomized, and will increase over time | • Develop visualizations of capability-based dashboards<br>• Define evaluation frequency and thresholds for error detection<br>• Audit and log LLM use in a standardized fashion, e.g. using MedLog[24], extended to include capability family/families for the task |
| Not feasible to manually monitor all of these metrics for all models | • Implement existing automatic metrics and identify gaps therein<br>• Develop new automated metrics based on identified gaps<br>• New automated metrics may include Judge LLMs: generative models used to evaluate outputs of other LLMs | • Scalable, continuous, and low-cost oversight | • Need to "audit the auditor" via periodic human calibration | Recommend tiered model:<br>• Automated screening by Judge LLMs and existing automated metrics (high-frequency, low-cost)<br>• Human review of flagged cases (low-frequency, high-interpretability) |
| Some truly high risk implementations will merit their own individualized oversight, (e.g. models making treatment recommendations without a human-in-the-loop) | • Maintain risk-stratified evaluation of an emerging technology. High risk devices still need the appropriate clinical testing before being integrated and monitored<br>• For models that are integrated, there will be a risk threshold at which organizations decide they still need individualized monitoring, but that will be the minority of cases | • Human oversight of very high risk models | • Humans may miss errors due to automation bias, over-reliance, and de-skilling | • Work with clinicians to investigate the feasibility of llm-as-judge or other monitoring method for a tiered approach over time |
| Difficult to support the needs of diverse stakeholders (health system leaders, clinical experts, and technical personnel that are distributed across the organization) with a standardized set of metrics | • Identify key stakeholder groups; conduct participatory design sessions with diverse stakeholders to develop prototypes of monitoring dashboards | • Supports teams of health system leaders, clinical experts, and technical personnel that are distributed across the organization as they monitor and respond to model deterioration | • Varied levels of technical expertise and knowledge may limit communication and understanding of metrics<br>• While streamlined compared to task-based monitoring, rapidly expanding capabilities may require ongoing reassessments | • Identify specific data views and functionality required by different stakeholders<br>• Periodically re-evaluate monitoring needs with stakeholders as models advance |
| Identified performance degradation will need to be addressed | • Develop standardized approach for root cause analysis<br>• Develop methods for correcting LLM performance<br>• Create back-up strategies for critical LLM-mediated functions | • Enables resilient model ecosystem that is robust to failures<br>• LLM performance degradation will not always require model fine-tuning; rapid prompt engineering and agentic updates may solve the problem.<br>• Capability-based monitoring enables shared solutions across workflows | • Limited insight and control over vendor LLMs<br>• Increasingly complex agentic systems with tool use and retrieval complicates root cause identification and resolution<br>• The same fix may not always work for all tasks, increasing workload | • Create best practices for manual review of errors, prompt review, and agentic system review<br>• Maintain ongoing communication and collaboration with vendors<br>• Ensure failure is due to LLM itself and not the surrounding architecture, which may be less generalizable<br>• Establish and maintain a database of example inputs for all workflows to confirm shared failure mode and resolution<br>• Maintain shared database of errors and solutions |
| **Implications for organizational leaders** | | | | |
| Capability-based monitoring at the business unit-level provides control for business-unit leaders, but risks missing cross-cutting vulnerabilities | • Centralize capability-based monitoring | • Centralized monitoring by capability reduces monitoring burden across hundreds of use cases, and enables cross-context evaluation of shared operations, early detection of systemic weaknesses, and identification of edge cases or rare errors | • Centralization reduces customization of solutions for each business unit and reduces the overall responsiveness to business unit needs<br>• Capability-based monitoring appropriate for post-deployment monitoring is not a substitute for initial needs assessment and evaluation | • Build team and resources to centralize capability-based monitoring<br>• Identify specific data views and dashboard functionality required by business unit stakeholders<br>Continue to perform initial needs assessment and evaluation by model |
| Merely detecting performance degradation is not sufficient | • Identify who is accountable for diagnosing the root cause of degradation, and applying strategies to restore model performance<br>• Develop a set of methods for root cause diagnosis and for restoring model performance<br>• Identify who needs to be informed of model issues, including taking models offline | • Ensures that degradations in model performance will be addressed and estimated ROI will continue to be realized | • Limited insight and control over vendor LLMs<br>• Increasingly complex agentic systems with tool use and retrieval complicates root cause identification and resolution<br>• The same fix may not always work for all tasks, increasing workload | • Maintain ongoing communication and collaboration with vendors<br>• Review failures with business unit leaders to ensure comprehensive understanding of failures and fixes<br>• Establish collaborations with other institutions to share identified errors and resolutions |
| Use of LLMs can deskill healthcare workers, making it difficult to take LLMs offline when deterioration is detected | • Institute requirements that professionals regularly practice high-impact tasks without AI, to maintain proficiency | • Can enable early detection of AI-induced deskilling in high-expertise domains | • Tradeoffs between deskilling solutions that minimize deskilling and those that | • Institute requirements that professionals regularly practice mission critical tasks without AI, to maintain proficiency |
| | • Leverage simulation technology to maintain task proficiency | | impose additional time and effort demands | |
| Lack of clear regulations makes it difficult to determine monitoring metrics | • Integrate with regulatory and accreditation processes by partnering with government affairs teams to create awareness of government agencies/regulatory bodies iteratively developing governance policies | • Supports the iterative development of metrics based on changing regulations | • AI technology will continue to move faster than external regulations | • Craft internal governance principles and governance process in advance of regulations<br>• Continue to monitor external regulations to align internal process with new regulations |
| Speed of change in models makes it difficult to determine which capabilities should be monitored | • Assign responsibility for external environmental scanning for new model capabilities | • Supports the monitoring of new capabilities<br>• Related capabilities may help anticipate future needs and failure modes<br>• Potential for cross-institution collaboration to learn from others' experiences | • Automated evaluation metrics will lag behind capability emergence, requiring more intensive initial manual oversight<br>• Emerging capabilities may initial resemble more traditional task-based, single workstream monitoring which may require bespoke visualizations and metrics<br>• Potential increased computational resource requirements for new models and monitoring thereof create a bottleneck limited by the institutional infrastructure and/or cost | • Establish internal team to review the literature for new capabilities, monitoring methods, and solutions<br>• Institute best practices for integrating a new capability family into monitoring dashboard<br>• Create communication structure for developers, informatics, and clinical team members to report gaps in capabilities<br>• Maintain reporting pathways for ad hoc error detection and requirements for critical harm reporting<br>• Create strategy for prioritizing model assessments to manage computational/cost resources |
| Individuals may develop LLM implementations for private use and not report these to the organization for monitoring | • Develop pathways and incentives for reporting bespoke workflows to organization | • Supports monitoring of all models being used by organization members | • Ease of developing new LLM workflows complicates identification and tracking of all uses<br>• Need for additional resources to identify and integrate uses into centralized monitoring systems | • Reward formalization of models: Provide recognition and rewards for formalizing new models<br>• Increase benefits associated with formalization: Provide resources for integration of models into EMR system so can be part of everyday workflow |
| **Implications for professional associations** | | | | |
| Different sophistication in LLM monitoring across institutions | • Develop shared benchmarks and reference frameworks across institutions | • Supports unified registries, clearer accountability, and consistent safety reporting | • In-house technical expertise required<br>• LLM-extrinsic monitoring dimensions may be highly | • Encourage sharing of frameworks, benchmarking strategies, and other monitoring resources via publication, |
| | | | sensitive and unique to specific institutions | presentation, and funding opportunities<br>• Formalize working groups and special conferences/workshops for dissemination and training |
| Inconsistent safety reporting | • Collaborative “monitoring commons” for healthcare AI safety | • Supports unified registries, clearer accountability, and consistent safety reporting | • Institutions must commit to logging LLM use according to shared protocols and taxonomies | • Centralized nationwide database for reporting LLM issues |

Abbreviations: LLM: Large Language Model; ROI: Return on Investment.